\documentclass{article}

 \usepackage[dblblindworkshop, final]{neurips_2025}
\workshoptitle{Evaluating the Evolving LLM Lifecycle Benchmarks, Emergent Abilities, and Scaling}

\usepackage[utf8]{inputenc} 
\usepackage[T1]{fontenc}    
\usepackage{hyperref}       
\usepackage{url}            
\usepackage{booktabs}       
\usepackage{nicefrac}       
\usepackage{microtype}      
\usepackage{xcolor}         

\usepackage{amsmath,amsfonts,amssymb}

\usepackage{wrapfig}
\usepackage{graphicx}
\usepackage{caption}

\definecolor{darkblue}{rgb}{0, 0, 0.5}
\hypersetup{colorlinks=true, citecolor=darkblue, linkcolor=darkblue, urlcolor=darkblue}

\usepackage{color}
\definecolor{darkred}{rgb}{0.6,0.0,0.0}
\definecolor{darkgreen}{rgb}{0,0.50,0}
\definecolor{lightblue}{rgb}{0.0,0.42,0.91}
\definecolor{orange}{rgb}{0.99,0.48,0.13}
\definecolor{grass}{rgb}{0.18,0.80,0.18}
\definecolor{pink}{rgb}{0.97,0.15,0.45}

\usepackage{listings}

\lstdefinelanguage{PythonPlus}{
  morestring=[b]",
  morekeywords=[1]{,print,class,in,for,import,from,as,assert,nonlocal,with,yield,self,True,False,None,} 
  morekeywords=[2]{,__init__,__add__,__mul__,__div__,__sub__,__call__,__getitem__,__setitem__,__eq__,__ne__,__nonzero__,__rmul__,__radd__,__repr__,__str__,__get__,__truediv__,__pow__,__name__,__future__,__all__,}, 
  morekeywords=[3]{,DatasetEntry,Dataset,CoTMetadata,CoT,LLMRequest,LLMResponse,LLMConfig,LLM,}, 
  morekeywords=[4]{,Exception,NameError,IndexError,SyntaxError,TypeError,ValueError,OverflowError,ZeroDivisionError,}, 
  morekeywords=[5]{,Optional,List,DatasetType,LLMType,BaseModel,str,float,int,} 
}
\lstdefinestyle{colorEX}{
  basicstyle=\ttfamily,
  backgroundcolor=\color{white},
  commentstyle=\color{darkgreen}\slshape,
  keywordstyle=\color{blue}\bfseries\itshape,
  keywordstyle=[2]\color{blue}\bfseries,
  keywordstyle=[3]\color{grass},
  keywordstyle=[4]\color{red},
  keywordstyle=[5]\color{orange},
  stringstyle=\color{darkred},
  emphstyle=\color{pink}\underbar,
}

\title{FEval-TTC: Fair Evaluation Protocol for Test-Time Compute}

\author{%
  Pavel Rumiantsev\\
  McGill University\\
  Montreal, Canada\\
  \texttt{pavel.rumiantsev@mail.mcgill.ca}
  \And
  Soumyasundar Pal\\
  Huawei Noah’s Ark Lab\\
  Montreal, Canada\\
  \texttt{soumyasundar.pal3@huawei.com}\\
  \And
  Yingxue Zhang\\
  Huawei Noah’s Ark Lab\\
  Montreal, Canada\\
  \texttt{yingxue.zhang@huawei.com}\\
  \And
  Mark Coates\\
  McGill University\\
  Montreal, Canada\\
  \texttt{mark.coates@mcgill.ca}\\
}

\begin{document}

\maketitle
\vspace{-0.25em}
\begin{abstract}
The performance of Large Language Models (LLMs) and the associated dollar costs of API calls can fluctuate over time, potentially invalidating conclusions drawn in prior research.
To address this, we propose a \textit{\textbf{F}air \textbf{Eval}uation protocol for \textbf{T}est-\textbf{T}ime \textbf{C}ompute} (FEval-TTC), designed to ensure consistent assessment of test-time compute (TTC) methods, regardless of such fluctuations.
FEval-TTC focuses on the evaluation of TTC methods that utilize underlying Chains-of-Thought (CoT).
It supports evaluations across multiple LLMs on a diverse set of mathematical and commonsense reasoning datasets.
The few-shot prompting and answer extraction processes are standardized across datasets, reducing both time and monetary overhead for researchers.
Furthermore, we provide a cost modelling procedure that estimates both the token and dollar cost per query, facilitating equitable comparisons of prevalent TTC methods.
We open-source FEval-TTC for public use at \url{https://github.com/networkslab/feval_ttc}.
\end{abstract}

\begin{figure}[h]
    \centering
    \vspace{-1.5em}
    \includegraphics[width=0.75\linewidth,keepaspectratio]{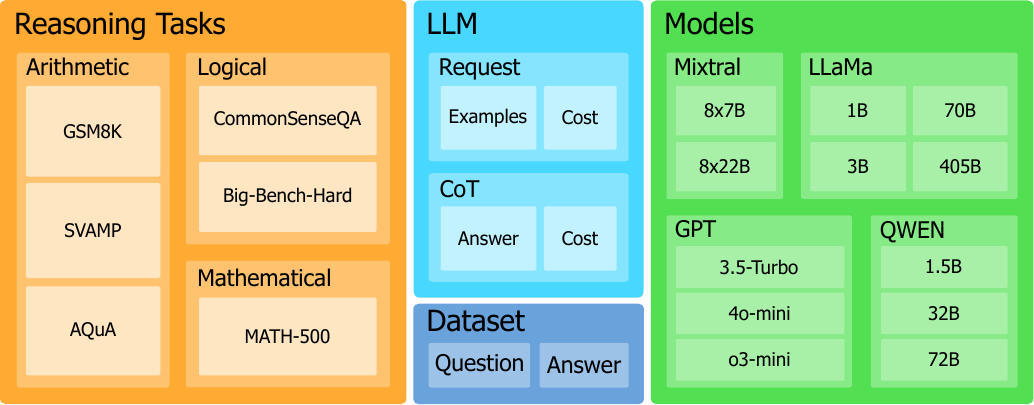}
    \caption{
    FEval-TTC comprises of three distinct groups of datasets, each consisting of question–answer pairs (see Section~\ref{sec:datasets}). 
    Each dataset is queried by multiple LLMs from different families with a standardized query format.
    We provide 40 sampled Chains-of-Thoughts (CoTs) with extracted answers, number of tokens, and the corresponding dollar cost of inference per question (see Section~\ref{sec:llms}).}
    \label{fig:ttbench_overview}
\end{figure}
\vspace{-1em}
\section{Introduction}
The emergence of System-2 thinking in Large Language Models~(LLMs)~\citep{ji2025test} has introduced a new paradigm that leverages inference-time computation to enhance reasoning capabilities~\citep{snell2025scaling}.
This paradigm involves allocating additional computational resources during inference, such as extended token generation, to improve performance on complex reasoning tasks~\citep{yang2025towards}.
The additional computation may originate from a single LLM~\citep{snell2025scaling} or from coordination among multiple LLMs~\citep{qi2025mutual}.
However, this increase in generation leads to substantial time and financial costs, posing practical challenges that hinder rapid experimentation and broader adoption.

In case a researcher is using an API, the monetary cost is primarily determined by commercial API usage fees, which are typically charged by API providers based on token usage. 
For self-hosted LLMs, monetary costs arise from the electricity consumed to operate GPUs. 
The time costs in both cases arise primarily from the latency between initiating a request and receiving the response, since the LLM generates tokens sequentially.
Concurrently, the LLM landscape evolves rapidly, with frequent model updates, new releases, and revisions to API pricing.
Such volatility can undermine the validity of prior research or create unfair advantages for newer methods if experimental setups do not carefully control for differences in model performance and cost.
Reusing results from published work without accounting for these changes can further exacerbate these issues, leading to inaccurate comparisons.

This \textbf{F}air \textbf{Eval}uation protocol for \textbf{T}est-\textbf{T}ime \textbf{C}ompute~(\textbf{FEval-TTC}) addresses these challenges by enabling researchers to substantially reduce both computational and time costs, while preserving fair and reproducible comparisons with prior work.
FEval-TTC includes a comprehensive set of pre-recorded model queries and responses, along with extracted answers and associated metadata.
For instance, applying self-consistency with 20 samples on the GSM8K dataset~\citep{cobbe2021gsm8k} using Mixtral 8×22B can take up to seven hours due to inference latency.
In contrast, FEval-TTC allows this evaluation to be completed in seconds by eliminating the need for live LLM calls.
We provide standardized requests and responses for sixteen datasets covering both commonsense and mathematical reasoning tasks.
Additionally, we introduce a unified cost model to ensure consistent and fair estimation of both query and response costs across different methods and models.

The uniqueness of FEval-TTC lies in the following features:
\begin{itemize}
    \item It supports several groups of reasoning tasks and multiple LLM model families.
    \item It is extensible in a trivial way to incorporate additional models, datasets, and prompting techniques.
    \item It ensures a fair comparison of test-time algorithms by using a standardized set of LLM responses and a unified monetary/token cost model.
    \item FEval-TTC significantly reduces the evaluation time and cost of common test-time inference methods by leveraging pre-recorded LLM responses instead of issuing live queries.
\end{itemize}

\vspace{-0.5em}
\section{FEval-TTC package overview}
\vspace{-0.75em}
\begin{lstlisting}[language=PythonPlus, caption={Example of an interaction with the \textit{FEval-TTC} package}, label=lst:ttbench_use_case,style=colorEX]
from feval_ttc import load, DatasetType, LLMType
    
dataset, [llm1,llm2] = load(DatasetType.SVAMP, \
                [LLMType.LLaMA3B32, LLMType.Qwen72B25])

for question_id, dataentry in dataset:
    print("Question: ", dataentry.question)
    print("True answer: ", dataentry.answer)
    llm1_response = llm1(question_id, N=20)
    print("1st CoT answer: ",  llm1_response.cots[0].answer)
    print("Token cost: ", llm1_response.cots[0].tokens)
    print("USD Cost: ", llm1_response.cots[0].dollar_cost)
\end{lstlisting}\vspace*{-1em}
This section provides the architectural overview of the FEval-TTC.
FEval-TTC is composed of two main parts: \textit{Dataset} module and \textit{LLM} module.
The Dataset module holds the list of questions and answers and an interface to iterate over them (see Section~\ref{sec:datasets}).
The LLM module stores multiple Chain-of-Thoughts~(CoTs) responses along with their extracted answers.
The design of both modules employs key-value dictionaries to facilitate seamless access to cached data.
The main package features an interface to load a Dataset module instance and a set of corresponding LLM module instances.
A usage example for research purposes is provided in Listing~\ref{lst:ttbench_use_case}.

\subsection{Dataset module}
\label{sec:datasets}
Dataset module instance contains a list of \textit{Dataentries} (see Listing~\ref{lst:ttbench_dataset}).
Each Dataentry includes a question and its ground-truth answer, collected from the corresponding datasets.
We did not change questions and answers, but the answer format was standardized across the package.
For each dataset, we provide a system prompt that was used to obtain LLM responses.

\begin{wrapfigure}{r}{0.48\textwidth}
\begin{minipage}{0.48\textwidth}
\vspace{-1.5em}
\begin{lstlisting}[language=PythonPlus, caption={Dataset module in FEval-TTC}, label=lst:ttbench_dataset,style=colorEX]
class DatasetEntry(BaseModel):
    answer: str
    question: str

class Dataset(BaseModel):
    data: List[DatasetEntry]
    datatype: DatasetType
    system_prompt: str
\end{lstlisting}
\vspace{-2em}
\end{minipage}
\end{wrapfigure}
FEval-TTC features datasets from three different reasoning categories: commonsense reasoning, arithmetic reasoning, and mathematical reasoning.
The \textbf{commonsense reasoning} group includes tasks designed to assess inference capabilities using commonsense knowledge, such as \textit{CommonSenseQA}~\citep{talmor2018commonsenseqa} and 11 \textit{BIG-Bench-Hard}~\citep{suzgun2023bbh} tasks.
The \textbf{arithmetic reasoning} group contains datasets such as \textit{GSM8K}~\citep{cobbe2021gsm8k}, \textit{SVAMP}~\citep{patel2021}, and \textit{AQuA}~\citep{ling2017}, which require basic calculation skills.
The \textbf{mathematical reasoning} category targets advanced problem-solving ability, represented by the \textit{MATH-500}~\citep{hendrycksmath2021}, which consists of competition-style mathematical questions requiring rigorous algebraic and geometric manipulation.

\vspace{-0.75em}
\subsection{LLM module}
\label{sec:llms}
\begin{wrapfigure}{r}{0.515\textwidth}
\vspace{-4em}
\begin{minipage}{0.515\textwidth}
\begin{lstlisting}[language=PythonPlus, caption={LLM modules in FEval-TTC}, label=lst:ttbench_llm,style=colorEX]
class CoTMetadata(BaseModel):
    dollar_cost: float
    tokens: int
    
class CoT(BaseModel):
    raw_text: str
    answer: Optional[str]
    metadata: CoTMetadata

class LLMRequest(BaseModel):
    raw_text: str
    dollar_cost: float
    tokens: int

class LLMResponse(BaseModel):
    cots: List[CoT]
    request: LLMRequest
    answers: List[str]

class LLMConfig(BaseModel):
    name: LLMType
    temperature: float
    max_tokens: int

class LLM(BaseModel):
    config: LLMConfig
    responses: List[LLMResponse]
\end{lstlisting}
\end{minipage}
\vspace{-2em}
\end{wrapfigure}

\textit{LLM} instance represents a real-world API, such as OpenAI\footnote{\url{https://platform.openai.com/docs/overview}}.
For each question, the instance returns an \textit{LLMResponse} object.
It provides access to a few-shot \textit{LLMRequest} prompt, which includes the official few-shot examples for a corresponding dataset, and to a set of \textit{CoT}.
Each CoT object consists of the raw API response and extracted answer. 
Note that not all CoTs contain answers that could be extracted; therefore, the answer field is set to \textit{None}, when such a failure occurs.
The answers extracted from CoTs are standardized across datasets.
The evaluation protocol is designed to be non-restrictive, allowing seamless integration with a researcher's existing methodology.
In practice, evaluating a test-time compute algorithm using FEval-TTC simply involves replacing live LLM API calls with provided responses.

We feature five common LLM families.
Each model is queried 40 times using a few-shot CoT prompt with standard few-shot examples.
The reasoning model o3-mini is queried 3 times using zero-shot instructions to save cost. Specifically, FEval-TTC includes CoTs from the following LLMs.\\
\textbf{LLaMA}: Llama 3.2-1B-Instruct, Llama 3.2-3B-Instruct, Llama 3.3-70B-Instruct, and Llama-3.1-405B-Instruct.\\ 
\textbf{QWEN}: Qwen2.5-1.5B-Instruct, Qwen2.5-32B-Instruct, and Qwen2.5-72B-Instruct.\\
\textbf{Deepseek}: Deepseek-V3.\\ 
\textbf{Mistral}: Mixtral-8x7B, and Mixtral-8x22B.\\
\textbf{GPT}: GPT 3.5 Turbo, GPT-4o-mini, and o3-mini~(reasoning).

\vspace{-0.5em}
\subsection{Dollar cost modelling}
\label{sec:cost_model}
FEval-TTC uses a unified monetary cost model to compute the dollar cost of an LLM response:
\begin{align}
\label{eq:dollar_cost}
DollarCost(\text{INP},\text{OUT}) = 10^{-6} \left( C_i\;Token(\text{INP}) + C_o\;Token(\text{OUT}) \right)\,,
\end{align}
where $C_i$ is the input processing cost of the model in USD per million tokens, $C_o$ is model's output cost in USD for generation of a million tokens, $Token(\text{INP})$ is the number of tokens in input prompt (from \textit{LLMRequest}) and $Token(\text{OUT})$ is the number of tokens generated by the LLM (from \textit{CoT}).
In our cost model, we assume that an LLM can be prompted once to sample multiple outputs, therefore, OUT may include multiple CoTs for a single input INP.

We adopt this simplified cost model to enable fair comparisons of LLM responses, independent of external factors such as the query date or caching strategies used.
We provide additional details in Appendix~\ref{app:cost_model}.

\vspace{-0.5em}
\section{Evaluation examples}
In order to demonstrate the use of our protocol, we present some examples of common Test-Time Compute methods evaluated on FEval-TTC.
Table~\ref{tab:common_ttc} and Figure~\ref{fig:aqua_sc_bon} show the results of Self-Consistency~\citep{wang2023selfconsistency} and Best-of-N~\citep{cobbe2021training} algorithms.
FEval-TTC also supports the evaluation of many existing training-free, adaptive self-consistency methods~\citep{aggarwal2023adapativesc,zhu2024path_consistency,wang2024make} for reducing the sampling cost of Self-Consistency.
Table~\ref{tab:multi_llm} and Figure~\ref{fig:math500_cascade} demonstrate the evaluation of multi-LLM (cascade) methods such as Mixture of Thoughts~\citep{yue2024large} and ModelSwitch~\citep{chen2025we}.
Other cascade approaches, such as FrugalGPT~\citep{aggarwal2024automix} and TREACLE~\citep{zhang2024treacle}, can also be evaluated using FEval-TTC. 

\begin{minipage}[c]{0.45\textwidth}
\centering
\captionof{table}{Accuracies of Self-Consistency~(SC) and Best-of-N~(BoN) with 20 CoTs with AQuA dataset.}
\label{tab:common_ttc}
    \small\begin{tabular}{lcc}
    \toprule
     Method & Mixtral 8x22B & Qwen 32B \\
    \midrule
     SC-20 & 0.787~(\$1.76) & 0.870~(\$0.24)  \\
     BoN-20 & 0.606~(\$1.76) & 0.870~(\$0.24) \\
    \bottomrule
    \end{tabular}
\end{minipage}\hfill%
\begin{minipage}[c]{0.45\textwidth}
\centering
\captionof{table}{Accuracies of Mixture of Thoughts~(MoT) and ModelSwitch~(MS) with LLaMA-70B and GPT-4o-mini.}
\label{tab:multi_llm}
    \small\begin{tabular}{lcc}
    \toprule
     Method & Ruin names & GSM8k \\
    \midrule
     MoT & 0.924~(\$0.44) & 0.960~(\$2.21)  \\
     MS & 0.916~(\$0.13) & 0.961~(\$0.64) \\
    \bottomrule
    \end{tabular}
\end{minipage}
\begin{minipage}[c]{0.45\textwidth}
\centering
    \includegraphics[width=\linewidth]{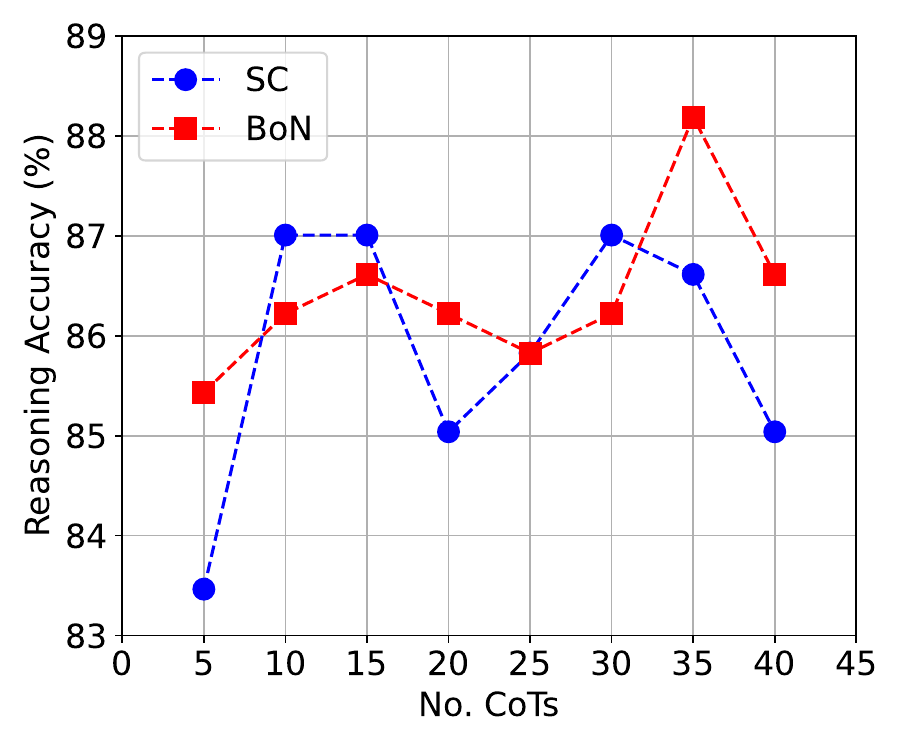}
    \captionof{figure}{Evaluation of CoT+SC and Best-of-N algorithms on AQuA dataset using Qwen 32B for varying number of CoTs.} 
    \label{fig:aqua_sc_bon}
\end{minipage}\hfill%
\begin{minipage}[c]{0.45\textwidth}
\centering
    \includegraphics[width=\linewidth]{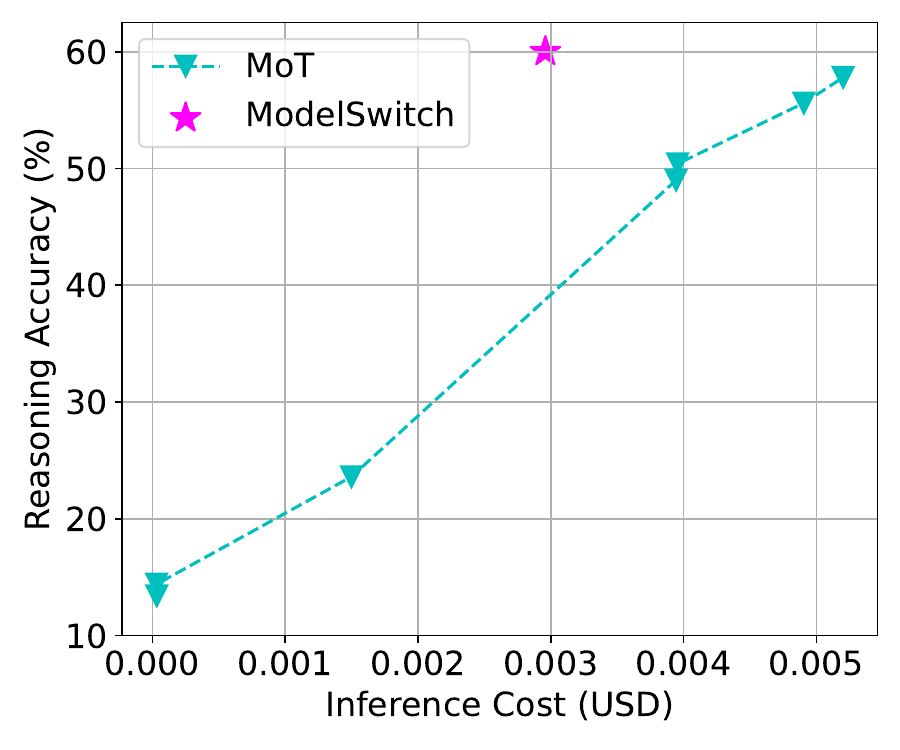}
    \captionof{figure}{Evaluation of MoT and ModelSwitch algorithms on MATH-500 dataset using Llama models for varying computational costs.} 
    \label{fig:math500_cascade}
\end{minipage}

\vspace{-1em}
\section{Conclusion}
We introduce FEval-TTC, an open-source framework for fast, fair, and low-cost evaluation of common test-time compute (TTC) methodologies.
By replacing LLM API calls with FEval-TTC API calls, researchers can reduce evaluation time from hours to seconds at negligible cost. 
Our unified cost model enables fair comparisons across methods, independent of API pricing fluctuations.
FEval-TTC facilitates the integration of new datasets and models through the application of standard prompting techniques.

\bibliography{colm2025_conference}
\bibliographystyle{colm2025_conference}

\appendix

\section{Unified cost model details}
\label{app:cost_model}

\begin{table}[h]
\centering
\caption{USD cost per million tokens for LLMs used in FEval-TTC. The costs are valid as of 02/06/2025.}
\label{tab:llm_prices}
\begin{tabular}{lcc}
\toprule
\textbf{LLM} & \textbf{Input Cost $C_i$ (\$/M tokens)} & \textbf{Output Cost $C_o$ (\$/M tokens)} \\
\midrule
LLaMA 3.2 1B-Instruct   & 0.005 & 0.01  \\
LLaMA 3.2 3B-Instruct   & 0.01  & 0.02  \\
LLaMA 3.3 70B-Instruct  & 0.13  & 0.40  \\
LLaMA 3.1 405B-Instruct & 1.00  & 3.00  \\
\midrule
Qwen 2.5 1B-Instruct  & 0.02 & 0.06 \\
Qwen 2.5 32B-Instruct  & 0.06 & 0.20 \\
Qwen 2.5 72B-Instruct  & 0.13 & 0.40 \\
\midrule
GPT 3.5-Turbo & 0.50 & 1.50 \\
GPT 4o-mini    & 0.15 & 0.60 \\
OpenAI o3-mini     & 1.10 & 4.40 \\
\midrule
Mixtral-8x7B-Instruct  & 0.08 & 0.24 \\
Mixtral-8x22B-Instruct & 0.40 &  1.20 \\
\midrule
DeepSeek-V3 & 0.50 & 1.50 \\
\bottomrule
\end{tabular}
\end{table}

FEval-TTC includes CoTs from various LLaMA, QWEN, Deepseek, Mistral, and GPT models.
We collected responses from the GPT model family using the OpenAI API\footnote{\url{https://openai.com/api}}.
OpenAI API prices are publicly available at \url{https://platform.openai.com/docs/pricing}.
We used a commercial API service\footnote{\url{https://docs.nebius.com/studio/inference/api}} to query other model families.
The price information for LLaMA, QWEN, Deepseek, and Mistral model families can be found at \url{https://nebius.com/prices-ai-studio}.
The detailed USD costs per million tokens for different LLMs are shown in Tables~\ref{tab:llm_prices}.
All costs were recorded as of June 2, 2025.

In our package, we provide access to both token counts and dollar cost.
Our unified model of dollar cost~\eqref{eq:dollar_cost} is proportional to the number of tokens.
Commercial API pricing is subject to change over time, typically at least once per year.
By fixing the dollar cost model in our protocol, we ensure that cost comparisons remain consistent and are unaffected by such pricing changes.
This design guarantees that comparisons between methods yield stable and fair conclusions, independent of future modifications to commercial API pricing policies.

\section{Licensing}
\label{app:licenses}

The terms of use for OpenAI API~\footnote{\url{https://docs.studio.nebius.com/legal/terms-of-service\#10-intellectual-property}\\``You hold exclusive ownership of all rights, titles, and interests (including intellectual property rights) to Your Inputs'' \& ``We do not claim any rights to Inputs and Outputs''} and Nebius API~\footnote{\url{https://openai.com/policies/row-terms-of-use/\#content}\\``As between you and OpenAI, and to the extent permitted by applicable law, you (a) retain your ownership rights in Input and (b) own the Output. We hereby assign to you all our right, title, and interest, if any, in and to Output.''} services grant the users full ownership of the LLM inputs and outputs provided by the API.
We distribute the collection of LLM inputs and CoT outputs under the Open Database License.
We grant rights of distribution, utilization, modification, and extension of the collection under the condition of a copyright notice.

Our Python package includes questions and ground truth answers for six datasets (including \textit{causal judgment, date understanding, disambiguationQA, formal fallacies, geometric shapes, movie recommendation, penguins, ruin names, snarks, sports, and temporal sequences} tasks of Big-Bench-Hard).
These datasets are provided for the convenience of the users.
We do not claim any ownership rights over the datasets included in the FEval-TTC package.
These datasets are independent assets distributed under the following licenses:
\begin{itemize}
    \item \textit{CommonSenseQA}~\citep{talmor2018commonsenseqa} under an MIT license
    \item \textit{Big-Bench-Hard}~\citep{suzgun2023bbh} under the MIT license
    \item \textit{GSM8K}~\citep{cobbe2021gsm8k} under the MIT license
    \item \textit{SVAMP}~\citep{patel2021} under the MIT license
    \item \textit{AQuA}~\citep{ling2017} under an Apache License, Version 2.0
    \item \textit{MATH-500}~\citep{hendrycksmath2021} under the MIT license
\end{itemize}

\end{document}